%% file: main.tex
\documentclass[10pt,twocolumn,letterpaper]{article}

\usepackage{iccv}
\usepackage{times}
\usepackage{epsfig}
\usepackage{graphicx}
\usepackage{amsmath}
\usepackage{amssymb}
\usepackage{amsmath}
\usepackage{bm}
\usepackage{enumitem}
\usepackage{booktabs}
\usepackage{parnotes}
\usepackage{caption}
\usepackage{pdfpages}
\usepackage{graphics,amssymb,amsmath,epsfig,color}
\usepackage{graphicx}
\usepackage[symbol]{footmisc}

\newcommand{\ours}{Background Prompting}
\newcommand{\supplementary}{{Supp.Mat.}}

\usepackage{array}
\newcolumntype{H}{>{\setbox0=\hbox\bgroup}c<{\egroup}@{}}

\usepackage[pagebackref=true,breaklinks=true,letterpaper=true,colorlinks,bookmarks=false]{hyperref}

\iccvfinalcopy %

\ificcvfinal\fi

\begin{document}

\title{Background Prompting for Improved Object Depth}
\author{
Manel Baradad\textsuperscript{1,*}
\and
Yuanzhen Li\textsuperscript{2}
\and
Forrester Cole\textsuperscript{2}
\and
Michael Rubinstein\textsuperscript{2}
\and
Antonio Torralba\textsuperscript{1}
\and
William T. Freeman\textsuperscript{1,2}
\and
Varun Jampani\textsuperscript{2}
\and
\textsuperscript{1} Massachusetts Institute of Technology \hspace{0.3in}   
\textsuperscript{2} Google Research
}

\makeatletter
\let\@oldmaketitle\@maketitle %
\renewcommand{\@maketitle}{\@oldmaketitle %
    \centering
    \vspace{-.2in}
    \includegraphics[width=\linewidth]{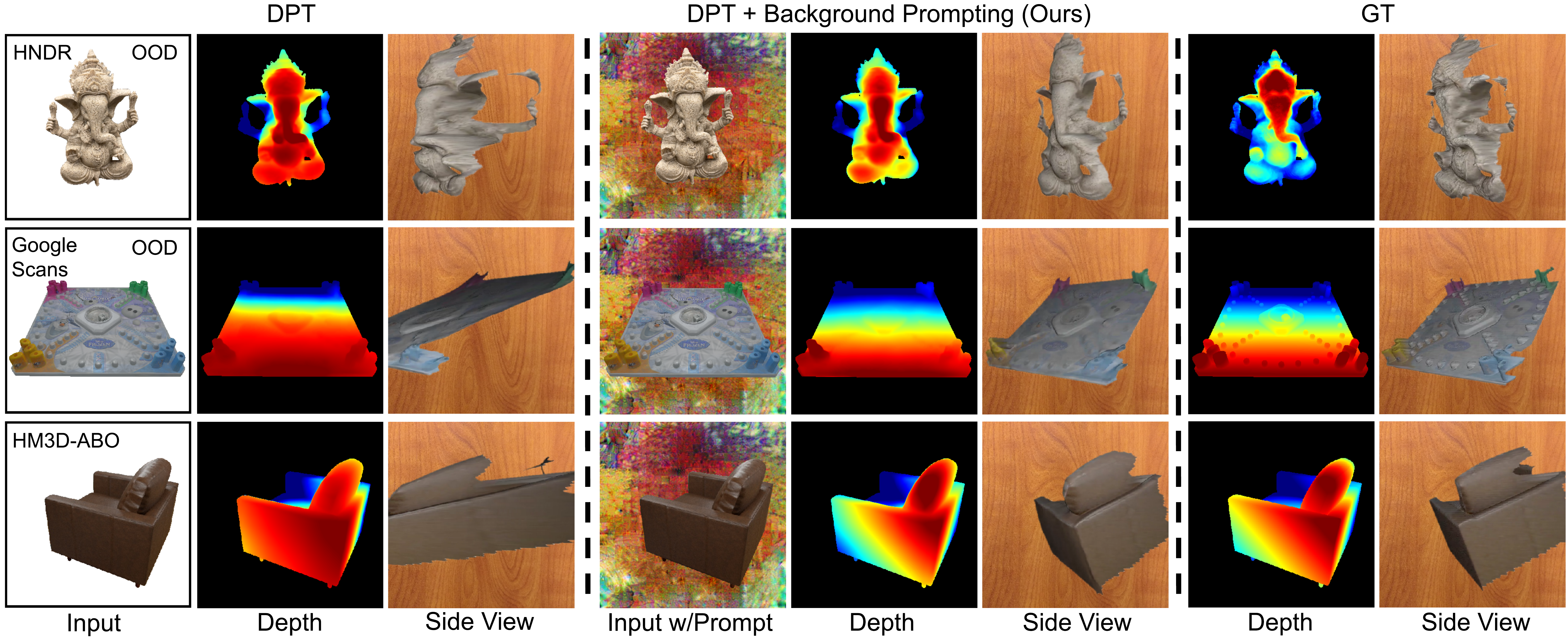}
    \captionof{figure}{\textbf{Improving object depth with background prompting}. Recovered depth and 3D from a single image using state-of-the-art DPT \cite{dpt} (left), Ours (DPT + \ours{}) (middle), compared to ground truth (right). Our method replaces the background pixels of the input with a learned background prompt, leading to improved depth. Prompts are trained with a small synthetic dataset (ABO), and yet perform well on held-out realistic datasets. Examples from HNDR (Out-of-Distribution), GoogleScans (OOD), and HM3D-ABO (In-distribution) are shown here.}
    \label{fig:teaser_results}
    \vspace{0.35in}}%
\makeatother
\maketitle
\ificcvfinal\thispagestyle{empty}\fi

\begin{abstract}
\vspace{-4mm}
Estimating the depth of objects from a single image is a valuable task for many vision, robotics, and graphics applications. However, current methods often fail to produce accurate depth for objects in diverse scenes.
In this work, we propose a simple yet effective \ours{} strategy that adapts the input object image with a learned background. 
We learn the background prompts only using small-scale synthetic object datasets. %
To infer object depth on a real image, we place the segmented object into the learned background prompt and run off-the-shelf depth networks.
\ours{} helps the depth networks focus on the foreground object, as they are made invariant to background variations. Moreover, \ours{} minimizes the domain gap between synthetic and real object images, leading to better sim2real generalization than simple finetuning.
Results on multiple synthetic and real datasets demonstrate consistent improvements in real object depths for a variety of existing depth networks. Code and optimized background prompts can be found at: \url{https://mbaradad.github.io/depth_prompt}.
\vspace{-4mm}
\end{abstract}

\input{01_introduction.tex}

\input{02_related_works.tex}
\input{03_method.tex}
\input{04_experiments.tex}

\input{05_analysis.tex}

\vspace{0.2mm}
\noindent \textbf{Limitations}.
We learn our BG prompting using full objects centered on the canvas. Although this can be easily corrected given the foreground mask by simulating a camera rotation and zoom, our method does not perform well if an object is only partially visible. This failure mode is more severe in the case of using a single background, as can be seen from the performance on the Nerfies dataset, which especially benefits from the Prompting network (see DPT results in Table~\ref{tab:out_of_distribution}). 

Experiments show that the performance of prompting is limited by the general performance of the original depth predictor. If the original network has been trained on small data sources and exhibits severe failure modes, prompting will likely be unable to recover from them. Shallow layer finetuning (and prompting at the input is an extreme case of it), may not be able to recover from considerable input and output distribution shifts \cite{surgical_finetuning}.

\section{Conclusion}
In this work, we propose a novel background prompting strategy to boost the object depth predictions from pre-trained depth networks. We find this approach to be quite training data efficient requiring only a small number of synthetic renderings. Our approach is also agnostic to the network architecture, showing consistent depth improvements across different depth models. In addition, we show good sim2real generalization with extensive experimental results on multiple datasets.

{\small
\bibliographystyle{ieee_fullname}
\bibliography{main}
}

\onecolumn

\pagebreak
\setcounter{equation}{0}
\setcounter{figure}{0}
\setcounter{table}{0}
\setcounter{page}{1}
\setcounter{section}{0}
\makeatletter
\renewcommand{\theequation}{S\arabic{equation}}
\renewcommand{\thefigure}{S\arabic{figure}}
\renewcommand{\thetable}{S\arabic{table}}

\include{supp_mat}

\end{document}

%% file: 01_introduction.tex
\section{Introduction}
\label{sec:intro}

\begin{figure}
\begin{center}
\includegraphics[width=1\linewidth]{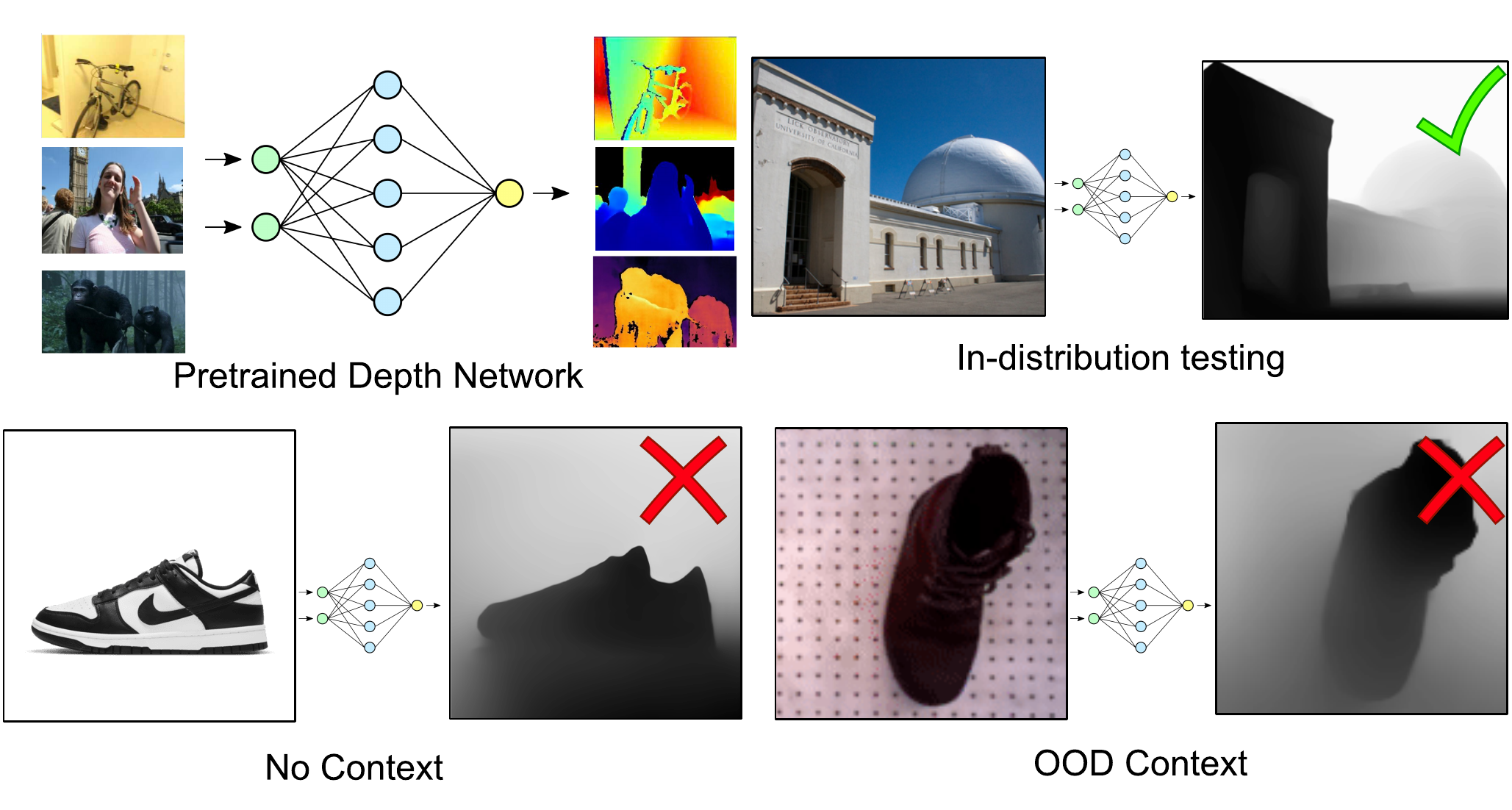}
\end{center}
\vspace{-0.2in}
\caption{\textbf{Poor object depths}. Off-the-shelf networks perform well when tested on in-distribution scenes but perform poorly on out-of-distribution data. This is exacerbated for object-centric images, where there is no context, or the context is out of distribution. We tackle this problem by learning backgrounds that adapt the input for depth networks to perform well in this setting. }
\vspace{-0.2in}
\label{fig:motivation}
\end{figure}

\footnotetext{
*Work done while interning at Google Research
}
Objects are central in our visual world, and obtaining good monocular depth for objects has a wide range of applications in vision (e.g.\ 3D reconstruction), robotics (e.g.\ object grasping), and graphics (e.g.\ relighting, defocus, photo-editing).
 Although recent neural networks~\cite{omnidata, dpt, midas,leres} have achieved impressive results on monocular depth estimation for various scenes (see Figure~\ref{fig:motivation} for examples), they often fail to capture accurate depth variations within the objects. Figure~\ref{fig:teaser_results} illustrates this by comparing the object depth estimations from a state-of-the-art method (DPT~\cite{dpt}) with ground truth. We can observe that estimated depths are inconsistent and unrealistic within the objects.

There are two main challenges for estimating object depth from a single image: 1)  Existing depth networks are trained on datasets that contain mostly scenes~\cite{scannet, mannequin, megadepth, midas},  as they are easier to obtain in the wild without requiring specialized lab setups. These networks learn to estimate the depth of different regions of the  scene, but not to produce accurate depths within each independent object. 2) The background of the image affects the performance for the object. If the object is isolated from its background (e.g.\ a white background) or if the background is unfamiliar to the network (e.g.\ a robotics setup), the input image becomes out-of-distribution for the network. This leads to unreliable object depth predictions, as shown in Figure~\ref{fig:motivation} using the state-of-the-art DPT depth network~\cite{dpt}.

A naive remedy for these challenges is to obtain large-scale real-world object-centric depth datasets, which are typically hard to obtain at scale. Even though synthetic yet realistic 3D object datasets~\cite{abo_dataset,google_scans} are readily available, it is non-trivial and tedious to place these objects in realistic and diverse background contexts to generalize the learned depth networks to real-world object images.
Works like~\cite{omnidata} tackle this problem by training on simulations of object-centric depths, but the quality of the synthetic images limits their transferability.

We introduce \emph{\ours{}}, a simple but effective technique using a learned background to improve the object depth predictions of existing depth networks. We first extract the object from the input image and put it on our learned background, replacing the original background with our learned background prompt. Then, we feed the resulting image with our learned background to the pre-trained depth network, obtaining improved object depths. We rely on existing foreground segmentation networks that are easy to use and work for any object category~\cite{rembg,dis_remove_background}.

Since real-world object datasets are difficult to collect, we use existing synthetic object datasets~\cite{abo_dataset} to learn our prompting backgrounds. Specifically, we train using object depth datasets with rendered objects in different views and their corresponding object depths. Then, we learn the background prompts via back-propagation through the frozen pre-trained depth network, with loss computed only on object pixels.
We experiment with two different background prompt parameterizations. One is directly optimizing a background RGB image that is agnostic to input. That is, the same learned background can be used for all objects (unconditional background prompting).
Another strategy is using a lightweight network that takes an object mask as input and estimates the background to be used for depth inference (conditional background prompting).

We refer to our background learning as `prompting' following the recent works that propose similar input adaptation strategies~\cite{hyojin_visual_prompting,visual_prompt_tuning}. To our knowledge, this is the first work that learns to prompt neural networks with learned backgrounds.
Our \ours{} technique has several favorable properties:
\begin{itemize}[leftmargin=*]
\vspace{-2mm}
\setlength{\itemsep}{0pt}
    \item \textbf{Sim2Real transfer}. By eliminating the background context of objects in training and inference, we effectively reduce the domain gap between simulated and real data, improving performance for images of real-world objects. Our method matches or surpasses finetuning in the out-of-distribution scenarios we examined, with additional benefits as follows.

    \item \textbf{Data efficient}. We only need a small dataset to learn background prompts, as the number of learnable parameters is orders of magnitude smaller than those in typical depth networks.
    \item \textbf{Easy use of synthetic training data}. Since we only need foreground object depth values during training, we can directly use synthetic 3D object datasets without any need for modeling realistic background context to place the synthetic objects. Realistic background modeling is non-trivial and often requires another background synthetic dataset, limiting the variations in the rendered dataset.
    \item \textbf{Network agnostic}. Since we only learn to modify the input
image, \ours{} is agnostic to network architecture. We show results with
both convolutional and transformer-based depth networks.
    \item \textbf{Reuse of existing networks}. \ours{} allows direct reuse
of pre-trained depth network weights without modifying any parameters or computation. This helps in preserving the depth prior learned by the
state-of-the-art networks.
    \item \textbf{Repurposing disparity networks for depth.} Most monocular
\emph{depth} networks only predict the scale and shift invariant disparity, i.e.\
the predicted disparities need to be shifted by an unknown factor before
being inverted to recover depth. Since we learn prompting using synthetic
objects, we can prompt the network to directly predict depths instead of
shift-invariant disparities.

\end{itemize}

Extensive experiments on synthetic and real-world datasets and using multiple network architectures demonstrate that using \ours{} results in consistent and reliable object depth improvements, both qualitatively and quantitatively. Code and pretrained models will be made available upon acceptance.

%% file: 02_related_works.tex
\section{Related work}
\label{sec:related_work}

\vspace{1mm}
\noindent \textbf{Depth from single image.}
Since the pioneering work of~\cite{eigen_depth}, depth prediction from a single image has seen consistent progress with multiple innovations, including novel losses~\cite{adabins, dorn, virtual_normal} and novel architectures~\cite{binsformer, plane_rcnn, dpt}. State-of-the-art methods now achieve outstanding performance when tested in-distribution on standard benchmarks like KITTI~\cite{kitti} or NYU~\cite{NYU}.

Given this progress, recent work has shifted the focus on generalization to diverse data sources. A key insight has been to train on diverse datasets, each providing different types of images and supervisory signals. To unify the different supervisory signals, works like MiDaS~\cite{midas} and LeReS~\cite{leres} penalize disparity up to an unknown additive constant and scale factor instead of penalizing metric depth. Using the scale-and-shift-invariant loss~\cite{scale_and_shif_invariant_loss}, it is possible to train jointly on diverse datasets with diverse ground truth, ranging from LIDAR depth to stereoscopic disparity obtained from 3D movies. While generalization to novel samples improves substantially using this loss, the unknown shift in disparity is often not easy to recover, making reconstructing metric depth from these systems difficult.

On the other hand, unsupervised approaches~\cite{digging_unsupervised_depth, sfmlearner_without_focal_length,unsupervised_dynamic_scenes,sfmlearner_with_radial_distortion,sfmlearner} allow training without depth ground-truth. Despite their potential to train at scale, unsupervised networks are not widely available and tend to perform worse than state-of-the-art supervised alternatives on novel data sources.

\vspace{1mm}
\noindent \textbf{Reusing depth networks.}
Recently, the work of~\cite{boosting_midas} proposed improving depth prediction using several forward passes of a depth network. The main insight of the method is that CNNs have a limited receptive field and produce different outputs depending on the input resolution. Using different resolutions, it is possible to obtain complementary coarse and fine-grained predictions, which then can be merged to outperform the original prediction. However, it is unclear how to adapt this approach to transformer-based architectures, which are the current state-of-the-art backbones~\cite{omnidata,dpt}, as their receptive field is the full image by design.  

Like our work,~\cite{adobe_repurpose} proposes using a foreground predictor to refine depth estimates. The technique substantially improves depth prediction along occlusion boundaries of the foreground and proposes a method to merge foreground and background depths. Unlike our approach, the technique requires retraining the depth prediction network.

\vspace{1mm}
\noindent \textbf{Network adaptation and visual prompting.} Large pretrained models are commonly adapted for downstream tasks, as they are too expensive to train from scratch.  \emph{Prompting} consists in adapting the network by making small changes to the input directly. Text prompts cannot be optimized with differentiable methods for discrete text inputs. Instead, prompting is done with non-differentiable techniques~\cite{autoprompt} or by modifying the embedding layers~\cite{prompt_tuning}.

Recent work has extended the prompting strategies to vision architectures. The pioneering work of~\cite{hyojin_visual_prompting} uses a learned border region on an image to adapt a pretrained classification network for a novel classification task. A recent work 
\cite{visual_prompt_tuning} studies a similar approach by appending new learned features to the input image, which changes the input shape and limits the approach to transformer-based architectures like DPT~\cite{dpt}. Another recent application of visual prompting~\cite{visual_prompting_via_inpaiting} seeks to repurpose an inpainting network to tackle specific vision tasks like segmentation, simply by constructing an inpainting task from several example model inputs and outputs and a masked reconstruction loss. Our method takes inspiration from these works but focuses on improving object depth prediction. 

%% file: 03_method.tex
\begin{figure*}
\begin{center}
\includegraphics[width=1\linewidth]{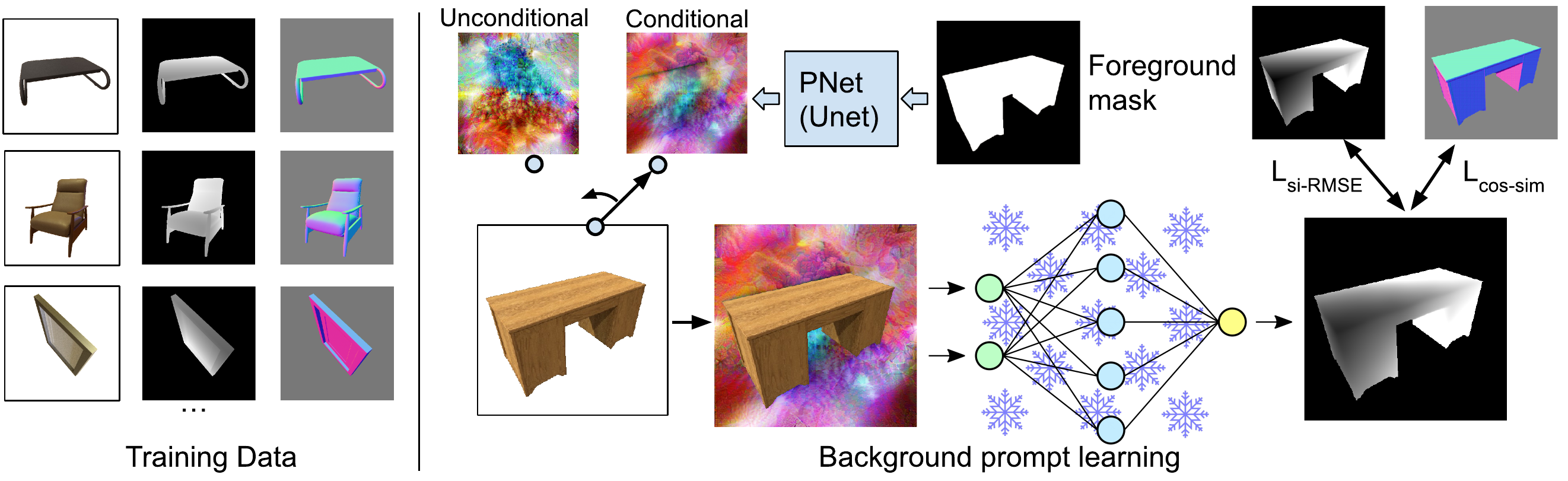}
\end{center}
\vspace{-0.2in}
\caption{\textbf{Learning background prompts.} We learn background prompts using rendered synthetic data, which improve predictions of off-the-shelf depth networks for object-centric depth. The depth network is kept frozen. We optimize the parameters of a single background prompt (unconditional) or a Unet that produces background prompts based on foreground masks (conditional), using standard losses for depth prediction.}
\vspace{-0.1in}
\label{fig:method_overview}
\end{figure*}

\section{Method}
\label{sec:method}

\noindent \textbf{Overview.}
Given a pre-trained network $\mathcal{D}$ that predicts depth (or disparity) up to scale and possibly an additive disparity factor (shift), we adapt it to improve object depth predictions. We do this by learning backgrounds instead of modifying the network parameters for the various efficiency reasons mentioned in Section~\ref{sec:intro}.
Formally, given an input RGB image $I \in \mathbb{R}^{n \times 3}$ with $n$ pixels,
we first cut-out the object $O \in \mathbb{R}^{n \times 3}$ using foreground segmentation $M \in \{0,1\}^{n \times 1}$.
Then we place it on our learned background $B \in \mathbb{R}^{n \times 3}$, resulting in a composite image $C \in \mathbb{R}^{n \times 3}$. Finally, we pass this composite image into the pre-trained and frozen depth network resulting
in improved depth estimates for object pixels. We use off-the-shelf high-quality foreground segmentators~\cite{rembg,dis_remove_background}, which are
readily available. In this work, we do not focus on homogenizing the improved object depth with the original background depth, and one could use existing techniques like~\cite{adobe_repurpose} for that purpose.

Following some recent works~\cite{hyojin_visual_prompting,visual_prompt_tuning} that learn to modify the network inputs, we call our background learning \emph{prompting}.
In contrast to~\cite{visual_prompt_tuning}, our background prompts are directly composited with the original images by replacing their background pixels and are not fed into the network as additional and specialized image tokens, which would restrict the applicability of the method to transformer-based architectures. Compared to~\cite{hyojin_visual_prompting}, our method does not require zooming out the image to add the prompt to the canvas as a border to the original image. Although this transformation is valid for some tasks invariant to zoom (e.g., classification or detection), zooming is not a valid input transformation for depth estimation, as depth prediction is not invariant to zooms.
The main research question in this work is: \textit{How to learn background prompts that can help boost the object depth estimates?}

\subsection{Background prompting}

Figure~\ref{fig:method_overview} illustrates the overview of our background prompting, where we use synthetic object datasets to learn the backgrounds. Using the synthetic object renderings, we learn background prompting by back-propagating through a depth network that is kept frozen. We propose different background prompting strategies which we discuss next.

\vspace{1mm}
\noindent \textbf{Unconditional and conditional background prompts}.
We want to learn background prompting on synthetic datasets while generalizing to real-world object images. As a result, we do not use input images directly to predict the backgrounds due to the domain gap between the synthetic and real images. Instead, we propose two learning strategies with no domain gap between synthetic and real.
One is to learn a single unconditional background that works for all the object images. This strategy generalizes well
to real-world images, as there is no input dependence on the image. The number of learned parameters is the number of background color values $3\times H\times W$. In the experiments, we see that this single universal background prompt works surprisingly well despite being simple and limited in the number of learnable parameters.

We also propose a more flexible strategy using a neural network $\mathcal{P}$ to predict the background prompt. 
This strategy is similar to using hyper networks~\cite{hypernetworks} to predict the model parameters. We call this conditioning network `Prompting network' or PNet for short.
Instead of using the object image directly as input to PNet, we use the object mask as input. Since the object masks exhibit a similar distribution in synthetic and real images, we find this strategy effective for generalizing PNet to real images. The number of learned parameters here is the number of PNet weights, typically much higher than the number of pixels in the unconditional background.

\vspace{1mm}
\noindent \textbf{Prompt parameterization.}
\begin{figure}
\begin{center}
\includegraphics[width=1\linewidth]{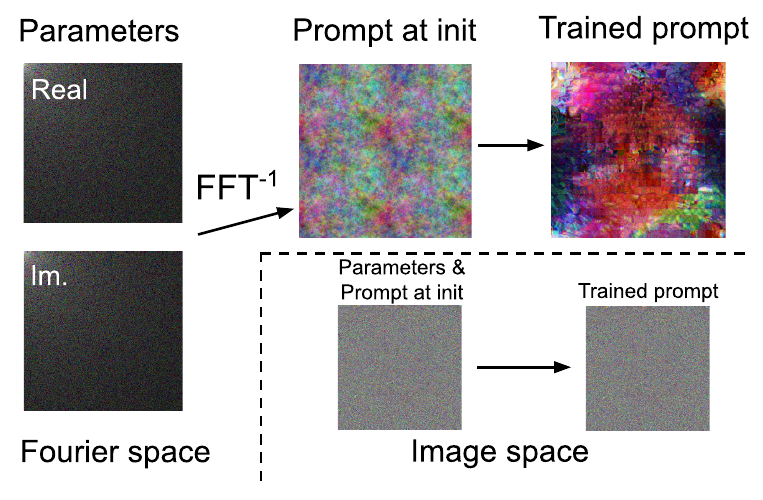}
\end{center}
\vspace{-0.2in}
\caption{\textbf{Learnt prompts and initializations.} We show prompts at initialization and after training, parameterized in Fourier space (our final model) and image space (which results in worse performance, as seen in the ablation study in Section~\ref{sec:ablation_study}). }
\label{fig:prompt_initialization}
\vspace{-0.2in}
\end{figure}
How we parameterize the backgrounds can have a significant effect on the results.
Learning background prompts in the Fourier space is more robust than directly predicting the raw RGB pixels
for multiple networks and datasets (see the ablation study in Section~\ref{sec:ablation_study}).

We follow the approach of~\cite{olah_feature_vis} and parameterize the background prompt by its real and imaginary parts $\hat{F} \in \mathbb{R}^{2\times 3\times H\times W}$, which are initialized following a $1/f$ power rule. This creates background prompts more similar to natural images  than Gaussian noise.
We convert the learned Fourier-domain background $F$ into RGB background $B$ using inverse FFT to create the composite RGB image to be passed into the network. In Figure~\ref{fig:prompt_initialization}, we show an example prompt at initialization and after training for both the Fourier space and the alternative of parameterizing directly in image space.  When using the Prompting network (PNet), we transform its input masks into the real and imaginary parts and treat the PNet outputs as the imaginary and real parts of the background prompt. Furthermore, we add a bias term to the output of the Prompting network, which in practice corresponds to adding an unconditional background prompt to the output of the Unet.

As different depth networks are trained using different input normalization schemes and expect the images to be normalized to these ranges, we also normalize the learned backgrounds to lie in the same range as the original inputs for each network. To do this, we pass the background prompt $B$ through a sigmoid layer $\sigma(.)$ and then apply the normalization function $\phi(.)$ of the original network being adapted. This makes the background prompts lie in the appropriate range for each network before they are composited with the foreground object images. Not doing so results in training instabilities, as the learned backgrounds saturate fast to ranges outside the expected input values of the networks.

\vspace{1mm}
\noindent \textbf{Compositing and inference}.
To combine the background prompt with the input image, we propose replacing non-object (background) pixels in the original images with the corresponding pixels in our learned background prompt.
Alternatively, the background prompt can be directly added to the input image (as is usually the case in adversarial attacks~\cite{explaining_adversarial_examples}). Still, this addition strategy performs worse (see ablations in Section~\ref{sec:ablation_study}). 

Formally, we compose the network input $C$ as:
\begin{align}
    C = M\phi(I)+(1-M)\phi(\sigma(B)),
\end{align}
where $M$ is the object mask; $I$ is the given input image; $B$ is the background prompt; $\phi$ and $\sigma$ are normalization and sigmoid functions respectively.

We then input the composed image into the depth network to get the object depth output. When the depth network is trained to predict disparity up to an unknown additive constant, as is the case of MiDaS~\cite{midas}, we use a fixed transform during both train and test time to transform output disparities $\hat{D} \in \mathbb{R}^{n \times 1}$ into depth $D \in \mathbb{R}^{n \times 1}$ without modifying the network. Following~\cite{photo_3d}, we use this transformation:

\begin{align}
\label{eq:depth_from_disparity}
D_p = max \left(\frac{\hat{D}_p - min(\hat{D})}{max(\hat{D}) - min(\hat{D})}, 0.05\right) ^{-1},
\end{align}
where $D_p$ and $\hat{D}_p$ denote the depth and disparity values at a pixel $p \in \{0,1,\cdots,n\}$.

\vspace{1mm}
\noindent \textbf{Synthetic data training and its advantages.}
Object-centric datasets are hard to collect, so we propose learning the prompts with synthetic data. This allows us to control for the diversity of the training data in terms of pose, viewpoints and, depth ranges, axes of variation that are hard to control with real data. Since the prompting only affects the input, we expect the network to preserve the capability of producing good geometry for real data if it has originally been trained on a large scale, which we verify through experiments.

\vspace{1mm}
\noindent \textbf{Training losses.}
We train the conditional and unconditional background prompts separately, using two standard losses for depth prediction:
One is the scale-invariant root-mean-square error (si-RMSE)~\cite{eigen_depth}:
\begin{align}
\label{eq:si_rmse}
L_{si-RMSE}(D, D^*) &= \min_{s}\left(  \frac{1}{|V|} \sum_{p \in V}(sD_p - D^*_p)^2 \right)^{1/2},
\end{align}
where $D^*$ denotes the ground-truth (GT) depth, $V$ represents the indices of the foreground object pixels, and $s$ is the scaling factor for the predicted depth that minimizes the loss with respect to GT.
The second loss is cosine similarity (cos-sim) on normals:
\begin{align}
\label{eq:cos_sim}
L_{cos-sim}(N, N^*)& =\frac{1}{|V|} \sum_{p \in V} \frac{1}{2} - \frac{1}{2} N_p \cdot N^*_p,
\end{align}
where $N$ and $N^*$ denote the estimated and GT unit normals, respectively. Normals are computed by reprojecting the depth using known intrinsics $K$ for the predicted and GT depths, respectively. The loss we use during training is the sum of~\ref{eq:si_rmse} and~\ref{eq:cos_sim}.

%% file: 04_experiments.tex
\section{Experiments}
\label{sec:experiments}

\noindent \textbf{Implementation details.}
We learn background prompting using a small dataset of synthetic objects for which ground-truth (GT) depths are available. 
We use the Amazon Berkeley Objects (ABO) dataset~\cite{abo_dataset}, which consists of $7.9k$ synthetic household objects. We use the renders available on the original dataset~\cite{abo_dataset} as well as the ABO-HM3D~\cite{hm3d_abo}  (HM3D), to have  a varied set of camera poses. The original dataset provides 91 renders per object, with viewpoints on the upper icosphere, while HM3D  consists of
the same synthetic objects as ABO, but is rendered with realistic poses and
illumination. We strip the backgrounds of the original images by replacing it with white color when evaluating off-the-shelf methods and finetuning. 

For object segmentation on images without foreground masks (HNDR and Nerfies), we use the rembg background segmenter~\cite{rembg}. We use the Unet architecture proposed in~\cite{pix2pix} for the Prompting network.
For all experiments, we train for $150k$ iterations, with a batch size of $8$ and vanilla SGD, with a learning rate of $5e-5$, annealed to $5e-5$ for the last $30k$ iterations. We use the same training schedule and loss for the baseline finetuning experiments but optimize the original network parameters instead.

\vspace{0.1mm}
\noindent \textbf{Datasets.}
We evaluate diverse synthetic and real-world dataset images.
To evaluate in-distribution performance, we use held-out ABO and HM3D  images.
To evaluate the generalization (out-of-distribution) performance, we evaluate
using 
1. Google Scans~\cite{google_scans}, which is a set of scanned real objects that we render from a diverse set of views; 2. HNDR~\cite{hndr} consists of high-quality images and depth maps for 10 objects; and 3. the public sequences of the Nerfies dataset~\cite{nerfies},  for which we train the original method and
render images and depths at the original camera poses.
We randomly sample 10K test images from these datasets when more than these are available.

\vspace{0.1mm}
\noindent \textbf{Metrics.}
We evaluate performance using scale-invariant RMSE (si-RMSE) and cosine similarity of normals, to measure how good is the estimated geometry, following similar works as~\cite{structdepth, midas}. To make results comparable across datasets and not penalize big objects more than small ones, we scale depths for each object to have a mean value of 5. We do this during
training and evaluation, and we fit predicted depths to ground truth (GT) when computing metrics following Eq.~\ref{eq:si_rmse}.

\vspace{0.1mm}
\noindent \textbf{Depth networks}.
We test our background prompting strategy using multiple state-of-the-art depth networks.
One is Omnidata~\cite{omnidata}, which is trained on synthetic renders of indoor scenes containing object-centric views.
We also experiment with MiDaS, with both convolutional (MiDaS-C)~\cite{midas} and transformer-based (DPT)~\cite{dpt} networks, which are trained using diverse datasets.
Lastly, we also experiment with LeReS~\cite{leres} with the Resnet-50 backbone, which is trained with a multi-objective loss to predict disparity or depth, depending on the available GT.

\vspace{0.1mm}
\noindent \textbf{Baselines and variations}.
We compare against the naive finetuning of the given depth networks using the same training data we use for background prompt learning.
We also compare with Boost-MiDaS~\cite{boosting_midas}, a technique to boost MiDaS depth on the MiDaS-C network,
but in this case, we do not adapt it with prompting as the method is not differentiable end-to-end. When evaluating Boost-MiDaS, we limit it to 1k random samples, due to computation constraints.
Within our background prompting strategies, we analyze both unconditional backgrounds (referred to as `1-BG')
and conditional backgrounds with prompting network (referred to as `PNet').

\begin{figure*}[t]
\begin{center}
\scalebox{0.88}{
\includegraphics[width=1\linewidth]{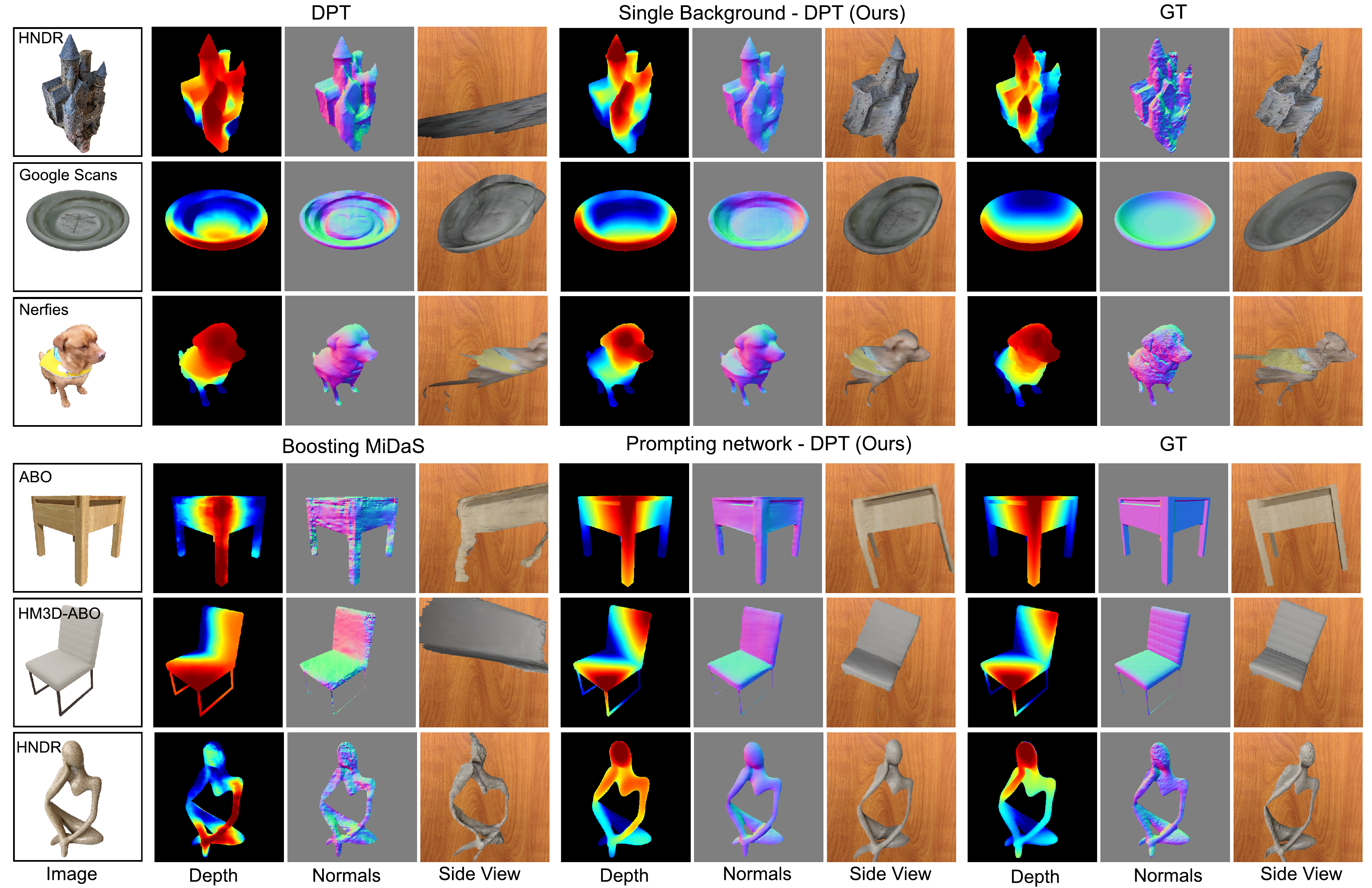}
}
\end{center}
\vspace{-0.20in}
\caption{\textbf{Qualitative results.} Depth, normals, and mesh (viewed from the side) for samples of each dataset, for our two methods and two baselines. Depths and meshes are scaled with Eq.~\ref{eq:si_rmse}. Side views are from the same position across methods. }
\label{fig:results_examples}
\vspace{-0.15in}
\end{figure*}

\begin{table}[]
\centering
\tabcolsep 4.4pt
\scalebox{1.0}{
\begin{tabular}{ll ll c ll}
\toprule
\textbf{Prompt } &
\textbf{Off-the-shelf} & 
\multicolumn{2}{c}{\textbf{ABO}} & &
\multicolumn{2}{c}{\textbf{HM3D}} \\
\cline{3-4} \cline{6-7} \\[-1em]

 & \textbf{network} &  si-R  $\downarrow$ & cos $\uparrow$ & & 
si-R $\downarrow$ & cos $\uparrow$ \\ \midrule

 & Boost MiDaS & $0.88$ & $0.65$ & & $0.85$ & $0.71$\\
 & MiDaS Conv & $1.10$ & $0.72$ & & $0.94$ & $0.79$\\
\textbf{None} & LeReS & $0.84$ &  $0.69$ & & $0.87$ & $0.72$\\
 & Omnidata & $0.36$ & $0.82$ & & $0.28$ & $0.83$\\  
 & DPT & $1.27$ & $0.81$ & & $1.09$ & $0.85$\\ \midrule

\textbf{1-BG}  & MiDaS Conv & $0.32$ & $0.87$ &  &$0.29$ & $0.89$\\  
\textbf{(Ours)} & LeReS & $0.35$ & $0.86$ & & $0.34$ & $0.88$\\  
\textbf{} & Omnidata & $0.24$ & $0.87$ & & $0.21$ & $0.88$\\  
& DPT & $0.27$ & $0.89$ &  &$0.25$ & $0.91$\\ \midrule 
 
 & MiDaS Conv & $0.32$ & $0.86$ & & $0.29$ & $0.89$\\  
\textbf{PNet} & LeReS & $0.32$ & $0.86$ & & $0.32$ & $0.88$\\  
\textbf{(Ours)} & Omnidata & $\mathbf{0.24}$ & $0.87$ &  &$\mathbf{0.21}$ & $0.88$\\
 & DPT & $0.26$ & $\mathbf{0.89}$ &  &$0.25$ & $\mathbf{0.91}$\\ \midrule

\end{tabular}
}
\caption{\textbf{In-distribution performance}. We compare the baseline off-the-shelf models against our proposed adaptation strategy, for the validation set of ABO and HM3D-ABO. Learning a single background and predicting it with the Prompting network substantially outperforms off-the-shelf networks. }
\vspace{-0.2in}
\label{tab:in_distribution}
\end{table}

\subsection{Results}

In Tables~\ref{tab:in_distribution} and~\ref{tab:out_of_distribution},  we report a subset of the evaluations for in-distribution and out-of-distribution settings.  The full set of experiments is available in the \supplementary, showing similar trends as the experiments highlighted in this section.  

\vspace{0.2mm}
\noindent \textbf{In-distribution performance}. Results in Table~\ref{tab:in_distribution} on ABO and
HM3D datasets demonstrate consistent performance improvements with respect to base depth networks.
The improvements are also consistent across different depth networks.
Some of them, like Omnidata have been trained on similar data sources, but still benefit substantially from our approach. 
For methods that predict disparity (like MiDaS and DPT), the learned backgrounds are able to adapt the network to predict depth by only changing the inputs and the fixed transformation described in Eq.~\ref{eq:depth_from_disparity}, while just applying the same transformation to the original network underperforms considerably. In addition, the improvements with background prompting are much bigger compared to those of strategies that reuse depth networks like Boost-MiDaS. In this setting PNet performs better or at least as well as the single background for all networks tested.

\begin{table}[]
\setlength\tabcolsep{4 pt}
\centering
\scalebox{1.0}{
\begin{tabular}{llccc}
\toprule
\textbf{Network } &
\textbf{Adapt} & 
\textbf{G.Scans} & 
\textbf{HNDR} &
\textbf{Nerfies} \\

\midrule
\textbf{MiDaS C} &\textbf{Ftune} &  $0.50$ & $0.58$  & $0.66$ \\
(Ours) & \textbf{1-BG} &  $\mathbf{0.50}$ & $0.57$ & $0.63$ \\
(Ours)& \textbf{PNet} & $0.51$ & $\mathbf{0.56}$  & $\mathbf{0.62}$ \\
\midrule

\textbf{Omnidata} &\textbf{Ftune} &  $0.37$ & $0.58$  & $0.61$  \\
(Ours) &\textbf{1-BG} &   $0.36$ & $0.57$  & $0.59$ \\
(Ours) & \textbf{PNet} & $\mathbf{0.35}$ & $\mathbf{0.55}$& $\mathbf{0.58}$\\
\midrule
\textbf{DPT} & \textbf{Ftune} & $0.70$  & $0.59$  & $0.62$ \\
(Ours) & \textbf{1-BG} & $\mathbf{0.43}$  & $0.60$ & $\mathbf{0.57}$ \\
(Ours) & \textbf{PNet} & $0.46$ & $\mathbf{0.58}$ & $0.78$\\
\midrule

\end{tabular}
}
\vspace{-0.05in}
\caption{\textbf{Out-of-distribution experiments.} si-RMSE ($\downarrow$) for finetuning the baselines against our adaptation strategies. Our method consistently performs better than finetuning on this metric, with PNet generally performing better than Single background.}
\vspace{-0.2in}
\label{tab:out_of_distribution}
\end{table}

\vspace{0.2mm}
\noindent \textbf{Out-of-distribution performance}
Table~\ref{tab:out_of_distribution} shows the results on datasets dissimilar to those seen during training: Google Scans, HNDR, and Nerfies. We compare the base models with finetuning and our two BG prompting strategies in this setting. Results show that 1-BG and PNet background prompting strategies generally perform better than alternatives across the depth networks. 
For the networks that originally performed the best (Omnidata and DPT), full finetuning on the synthetic data usually yields worse performance than our strategy regarding si-RMSE. In this case, ours achieves considerably better results than finetuning, as the network preserves the priors of the original Omnidata network, which are lost during finetuning.

When considering Single Background compared to PNet, results are mixed, PNet underperforming for some combinations of networks and datasets with respect to single background. We believe that this is because certain datasets like Nerfies have a distribution of masks that is different than the training one (e.g. not centered on the frame, which we do not account for during training nor testing).

\begin{table}[]
\setlength\tabcolsep{4 pt}
\centering
\begin{tabular}{llllll}
\toprule
\textbf{Method } &
\textbf{Ablation} &
\multicolumn{2}{c}{\textbf{ABO}} & 
\multicolumn{2}{c}{\textbf{HNDR}} \\
\cmidrule(lr){3-4}\cmidrule(lr){5-6} \\[-1.2em]
& &  si-R $\downarrow$ & cos $\uparrow$ &
si-R $\downarrow$ & cos $\uparrow$ \\ \midrule
\textbf{1-BG} & Additive  & $0.30$ & $0.87$ & $0.66$ & $0.84$ \\
 & Img space & $0.40$ & $0.86$ & $0.84$ & $0.83$ \\  

& Full & $0.26$ & $0.89$ & $0.62$ & $0.85$ \\
\midrule
\textbf{PNet} & Img space & $0.25$ & $0.89$ & $0.61$ & $0.84$ \\
&Input img & $0.21$ & $0.90$ & $0.60$ & $0.84$ \\
&No bias & $0.26$ & $0.89$ & $0.59$ & $0.85$ \\
&Full & $0.25$ & $0.89$ & $0.56$ & $0.84$ \\
\midrule
\end{tabular}
\vspace{-0.05in}
\caption{\textbf{Ablation studies}. Performance ablating components of our method trained with ABO. We test additive background instead of composting (Additive), training in images space instead of Fourier space (Img space), using images as input to the PNet instead of the foreground masks  (Input img), and removing the bias from the PNet (No bias).}
\label{tab:ablations}
\vspace{-0.2in}
\end{table}

\vspace{0.2mm}
\noindent \textbf{Ablation studies}.
\label{sec:ablation_study}
Table \ref{tab:ablations} shows results with the ablation of several design choices in our learning framework.
We report results on two representative datasets, ABO (in-distribution) and HNDR (out-of-distribution), with the results for all the datasets available in the \supplementary\ and showing similar trends. 
For the unconditional strategy (1-BG), we compare against 1) using an additive noise instead of replacing the background values (Additive) and 2) parameterizing the prompts on pixel space and initializing them with Gaussian noise (Img Space). Results in Table \ref{tab:ablations} show that both alternatives yield lower performance on both datasets, showing that adaptation performs worse without these two components. 
The visual prompts recovered for the two ablations can be seen in Figure~\ref{fig:learnt_prompts}, illustrating their failure modes. For the additive prompt, we observe that only the borders where the objects are never placed have high frequencies. As additive prompts can potentially corrupt object pixels, the prompting strategy learns to place the prompt at the borders.

We also ablate the spectrum prediction for the PNet and the additive bias term at the output, underperforming the complete proposed method. We also test using the foreground images instead of the masks as inputs to the prompting network. Although in-distribution performance is higher, the learned prompts do not transfer well to other datasets compared to foreground masks. This is expected, as the images are more informative than the masks, but it may lead to overfitting to the image details, which don't transfer to other datasets.

%% file: 05_analysis.tex
\begin{figure}
\begin{center}
\includegraphics[width=1\linewidth]{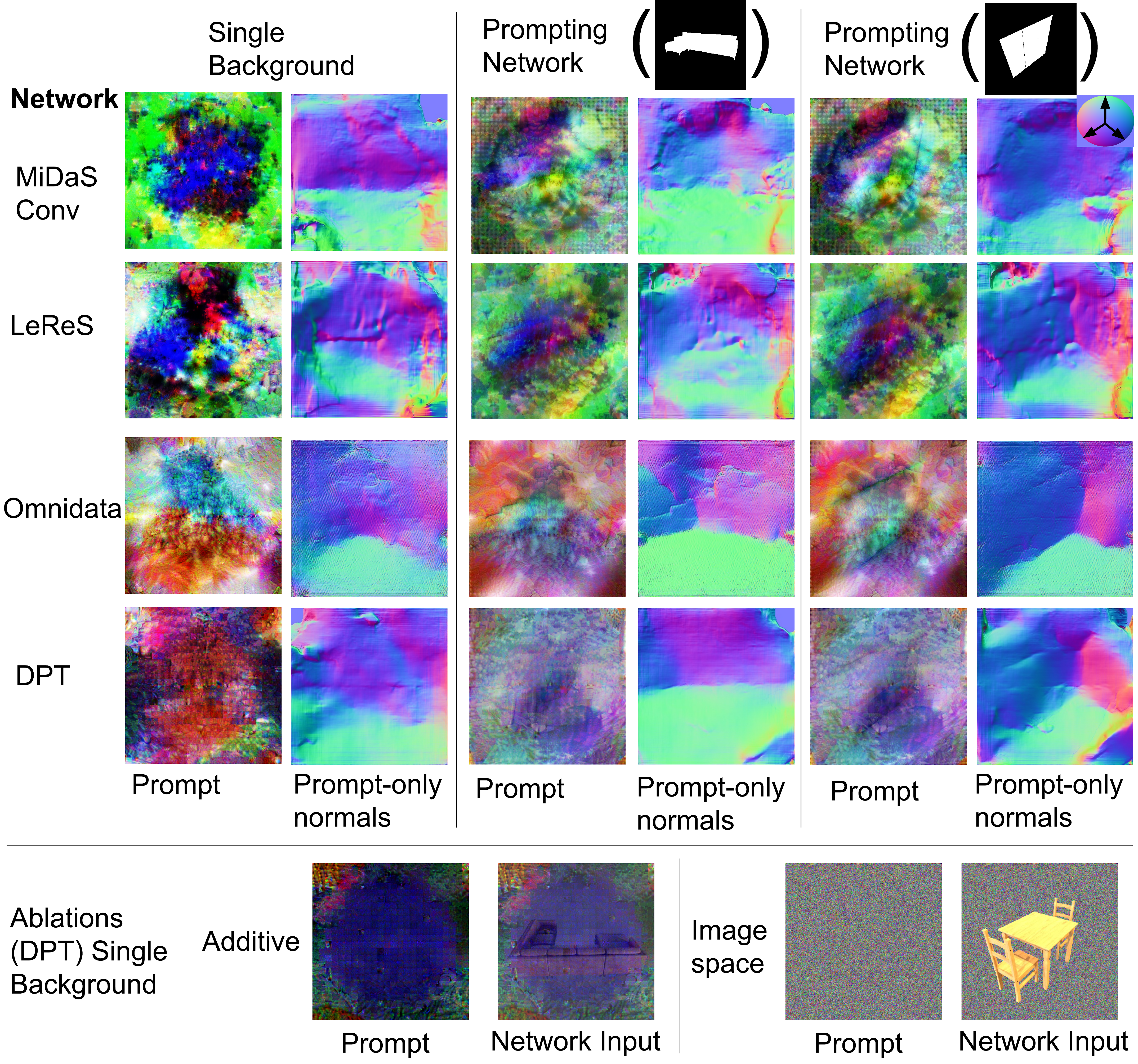}
\end{center}
\vspace{-0.2in}
\caption{\textbf{Learned backgrounds and predicted normal maps}. These are obtained by feeding the background prompts to the networks without the foreground object being inpainted. The last row shows the prompts learned for two ablations in Sec.\ \ref{sec:ablation_study}.}
\vspace{-0.2in}
\label{fig:learnt_prompts}
\end{figure}

\vspace{0.2mm}
\noindent \textbf{Visual results}.
Figure~\ref{fig:results_examples} shows sample object results of our method along with some baselines, with more visuals in Figure~\ref{fig:teaser_results}. 
Results show that our background prompting allows DPT networks to predict depths with more details closer to GT depths and normals. Novel views of the object highlight the improvement in depth with our prompting.
Additional results can be found in the \supplementary\ 

Compared with Boosting Midas \cite{boosting_midas}, our method can also correctly predict fine-grained details (like the legs of the chair and thin structures of the sculpture) while not introducing any unrealistic high-frequency details or inconsistent global depth.

In Figure \ref{fig:learnt_prompts}, we show the visuals of the learned background prompts for different networks and the corresponding normal maps obtained by passing only the prompt image into the network. Results show that without conditioning, the prompts produce a ground plane, a standard structure in most datasets, that depth networks tend to use to propagate depths \cite{how_do_nn_see_depth}. When conditioning with PNet, we see a similar behavior across networks: prompts adapt to produce a background depth that provides perspective cues that match the object shape, similar to a box aligned with the foreground object, 
as can be seen in the right-most columns.

%% file: supp_mat.tex
\begin{center}
    \Large{\@setfontsize\Large\@xviipt{22}}
      {\Large \bf Background Prompting for Improved Object Depth - Supplementary Material \par}
      \vspace*{12pt}
\end{center}

\section{Extended Experimental results}
\subsection{Full results}
\label{sec:intro}

In Table~\ref{tab:full_results} we show results for the full set of experiments described in Section 4 of the main paper.  Our strategy is able to adapt all methods to perform much better than the original methods without adaptation. Furthermore, our method tends to perform more robustly on si-RMSE than finetuning when tested out-of distribution for the best performing networks. This can be seen by comparing finetuning and PromptNet for Omnidata and DPT on out-of-distribution datasets (Google Scans, HNDR and Nerfies).

\begin{table*}[h]
\centering
\scalebox{0.8}{
\begin{tabular}{l l |ll| ll|ll|ll|ll}
\toprule
\textbf{Adapt} & \textbf{Pretrained} & 
\multicolumn{2}{c|}{\textbf{ABO}} & 
\multicolumn{2}{c|}{\textbf{HM3D}} & 
\multicolumn{2}{c|}{\textbf{G Scans}} & 
\multicolumn{2}{c|}{\textbf{HNDR}} & 
\multicolumn{2}{c}{\textbf{Nerfies}}\\
\textbf{Method} & \textbf{network} &
si-R $\downarrow$& cos$\uparrow$ &
si-R $\downarrow$& cos$\uparrow$ &
si-R $\downarrow$& cos$\uparrow$ &
si-R $\downarrow$& cos$\uparrow$ &
si-R $\downarrow$& cos$\uparrow$ \\ \midrule
 & Boosted MiDas & $0.88$ & $0.65$ & $0.85$ & $0.71$ & $1.09$ & $0.57$ & $0.77$ & $0.69$ & $0.86$ & $0.48$\\  
\textbf{None} & MiDaS Conv & $1.11$ & $0.73$ & $0.91$ & $0.80$ & $1.00$ & $0.71$ & $0.84$ & $0.80$ & $0.90$ & $0.62$\\  
 & LeReS & $0.84$ & $0.69$ & $0.87$ & $0.72$ & $0.77$ & $0.68$ & $0.89$ & $0.68$ & $0.89$ & $0.48$\\  
 & Omnidata & $0.36$ & $0.82$ & $0.28$ & $0.83$ & $0.35$ & $0.81$ & $0.60$ & $0.76$ & $0.59$ & $0.59$\\  
 & DPT & $1.27$ & $0.81$ & $1.09$ & $0.85$ & $0.93$ & $0.80$ & $1.32$ & $0.79$ & $1.02$ & $0.62$\\ \midrule 
 & MiDaS Conv & $0.12$ & $0.93$ & $0.09$ & $0.95$ & $0.50$ & $0.86$ & $0.58$ & $0.82$ & $0.66$ & $0.64$\\  
\textbf{Finetune} & LeReS & $0.12$ & $0.94$ & $0.10$ & $0.96$ & $0.41$ & $0.87$ & $0.64$ & $0.80$ & $0.72$ & $0.61$\\  
 & Omnidata & $0.12$ & $0.94$ & $0.10$ & $0.95$ & $0.37$ & $0.88$ & $0.58$ & $0.84$ & $0.61$ & $0.65$\\  
 & DPT & $0.12$ & $0.94$ & $0.10$ & $0.95$ & $0.70$ & $0.85$ & $0.59$ & $0.84$ & $0.62$ & $0.66$\\ \midrule 
 & MiDaS Conv & $0.32$ & $0.87$ & $0.29$ & $0.89$ & $0.50$ & $0.83$ & $0.57$ & $0.86$ & $0.63$ & $0.65$\\  
\textbf{Single} & LeReS & $0.35$ & $0.86$ & $0.34$ & $0.88$ & $0.47$ & $0.80$ & $0.62$ & $0.83$ & $0.75$ & $0.58$\\  
 & Omnidata & $0.24$ & $0.87$ & $0.21$ & $0.88$ & $0.36$ & $0.84$ & $0.57$ & $0.80$ & $0.59$ & $0.59$\\  
 & DPT & $0.27$ & $0.89$ & $0.25$ & $0.91$ & $0.43$ & $0.86$ & $0.60$ & $0.85$ & $0.57$ & $0.65$\\ \midrule 
 & MiDaS Conv & $0.32$ & $0.86$ & $0.29$ & $0.89$ & $0.51$ & $0.83$ & $0.56$ & $0.85$ & $0.62$ & $0.65$\\  
\textbf{HyperNet} & LeReS & $0.32$ & $0.86$ & $0.32$ & $0.88$ & $0.49$ & $0.81$ & $0.63$ & $0.83$ & $0.75$ & $0.58$\\  
 & Omnidata & $0.24$ & $0.87$ & $0.21$ & $0.88$ & $0.35$ & $0.84$ & $0.55$ & $0.79$ & $0.58$ & $0.59$\\  
 & DPT & $0.26$ & $0.89$ & $0.25$ & $0.91$ & $0.46$ & $0.86$ & $0.58$ & $0.84$ & $0.78$ & $0.64$\\ \midrule

\end{tabular}
}
\caption{Performance for off-the-shelf networks without adaptation method and with finetuning, and our proposed method with the same networks with a single learnt background and with the Prompting Network predicting the background. }
\label{tab:full_results}
\end{table*}

\subsection{Full Ablation results}
In Table~\ref{tab:full_ablation_results} we show the results with the same ablation studies as in Section 4.1 of the main paper, with the main conclusions of the study found in that section extending to the rest of the datasets.

\subsection{Background baselines}
In Table~\ref{tab:background_randomization}, we evaluate various composite methods that employ random backgrounds without the process of fine-tuning. The technique detailed in the main paper utilizes a white background (None) and has been shown to consistently surpass the effectiveness of other background types. These include the original background (Original), random Gaussian noise (Rand), and a randomly selected real image from the Places365 dataset (Places). As seen in the table, our method is substantially superior to fixed background types across all networks, even outperforming the original background.

\begin{table*}[h]
\centering
\begin{tabular}{l l |ll| ll|ll|ll|ll HH}
\toprule
\textbf{Adapt} & \textbf{Ablation} &
\multicolumn{2}{c|}{\textbf{ABO}} &
\multicolumn{2}{c|}{\textbf{HM3D}} &
\multicolumn{2}{c|}{\textbf{G Scans}} &
\multicolumn{2}{c|}{\textbf{HNDR}} &
\multicolumn{2}{c}{\textbf{Nerfies}} &
\multicolumn{2}{H}{\textbf{DTU}}  \\
\textbf{Method} & \textbf{} &
si-R $\downarrow$& cos$\uparrow$ &
si-R $\downarrow$& cos$\uparrow$ &
si-R $\downarrow$& cos$\uparrow$ &
si-R $\downarrow$& cos$\uparrow$ &
si-R $\downarrow$& cos$\uparrow$ &
si-R $\downarrow$& cos$\uparrow$  \\ \midrule

\textbf{Background} & Additive & $0.30$ & $0.87$ & $0.41$ & $0.86$ & $0.47$ & $0.85$ & $0.66$ & $0.84$ & $0.82$ & $0.62$ & $1.40$ & $0.64$\\  
 \textbf{Single} & Img space & $0.40$ & $0.86$ & $0.36$ & $0.89$ & $0.63$ & $0.83$ & $0.84$ & $0.83$ & $1.00$ & $0.65$ & $1.79$ & $0.68$\\  
 & None & $0.26$ & $0.89$ & $0.36$ & $0.90$ & $0.45$ & $0.86$ & $0.62$ & $0.85$ & $0.79$ & $0.63$ & $1.52$ & $0.69$ \\

\midrule 
\textbf{PromptNet} & No bias & $0.25$ & $0.89$ & $0.38$ & $0.89$ & $0.48$ & $0.86$ & $0.61$ & $0.84$ & $0.95$ & $0.62$ & $1.57$ & $0.71$\\  
\textbf{} & Img space & $0.21$ & $0.90$ & $0.37$ & $0.90$ & $0.52$ & $0.85$ & $0.60$ & $0.84$ & $0.86$ & $0.62$ & $1.52$ & $0.67$\\  
\textbf{} & Img input & $0.26$ & $0.89$ & $0.48$ & $0.89$ & $0.50$ & $0.86$ & $0.59$ & $0.85$ & $0.63$ & $0.64$ & $1.75$ & $0.69$\\  
 & None & $0.25$ & $0.89$ & $0.37$ & $0.90$ & $0.44$ & $0.86$ & $0.56$ & $0.84$ & $0.59$ & $0.65$ & $1.56$ & $0.69$\\

\end{tabular}
\caption{Performance for off-the-shelf networks with different ablated components for our method as described in Seciton 4.1 of the main paper and extending Table of the main paper.}
\label{tab:full_ablation_results}
\end{table*}

\begin{table*}[h]
\centering

\tabcolsep 4.4pt
\scalebox{0.85}{
\begin{tabular}{ll ll c ll}
\toprule
\textbf{Off-the-shelf} & 
\textbf{BG } &
\multicolumn{2}{c}{\textbf{ABO*}} & &
\multicolumn{2}{c}{\textbf{HNDR}} \\
\cline{3-4} \cline{6-7} \\[-1em]
\textbf{network} & &  si-R  $\downarrow$ & cos $\uparrow$ & & 
si-R $\downarrow$ & cos $\uparrow$ \\ \midrule

 & None& $1.11$ & $0.73$& & $0.84$ & $0.80$\\  
\textbf{MiDaS Conv} & Original& $1.15$ & $0.79$& & $1.06$ & $0.84$\\  
 & Random& $0.93$ & $0.73$& & $0.97$ & $0.81$\\ 
 & Places& $1.57$ & $0.77$& & $1.64$ & $0.82$\\  
 & Ours & $\mathbf{0.32}$ & $\mathbf{0.86}$ && $\mathbf{0.57}$ & $\mathbf{0.86}$  \\
\midrule 
 & None& $0.36$ & $0.82$& & $0.60$ & $0.76$\\  
\textbf{Omnidata} & Original& $0.62$ & $0.77$& & $0.60$ & $0.74$\\  
 & Random& $0.46$ & $0.80$& & $0.63$ & $0.72$\\ 
 & Places& $0.74$ & $0.72$& & $0.71$ & $0.72$\\  
 & Ours & $\mathbf{0.24}$ & $\mathbf{0.87}$  && $\mathbf{0.57}$ & $\mathbf{0.80}$\\
\midrule 
 & None& $1.27$ & $0.81$& & $1.32$ & $0.79$\\  
\textbf{DPT} & Original& $1.27$ & $0.83$& & $1.34$ & $0.82$\\  
 & Random& $1.99$ & $0.70$& & $2.61$ & $0.80$\\ 
 & Places& $1.55$ & $0.76$& & $1.66$ & $0.79$\\  
 & Ours & $\mathbf{0.26}$ & $\mathbf{0.89}$ && $\mathbf{0.60}$ & $\mathbf{0.85}$ \\
\midrule 
\end{tabular}
}%
\caption{Evaluation using different backgrounds. `None' (same as main paper) corresponds to white background, `Orig.'\ to the original background found in the dataset, `Places' to a random image in the Places365 dataset, and `Rand' to iid.\ uniform RGB noise. Ours corresponds to the single background approach.}
\label{tab:background_randomization}
\end{table*}

\section{Qualitative results} 
\begin{figure*}[h]
\begin{center}
\includegraphics[width=1\linewidth]{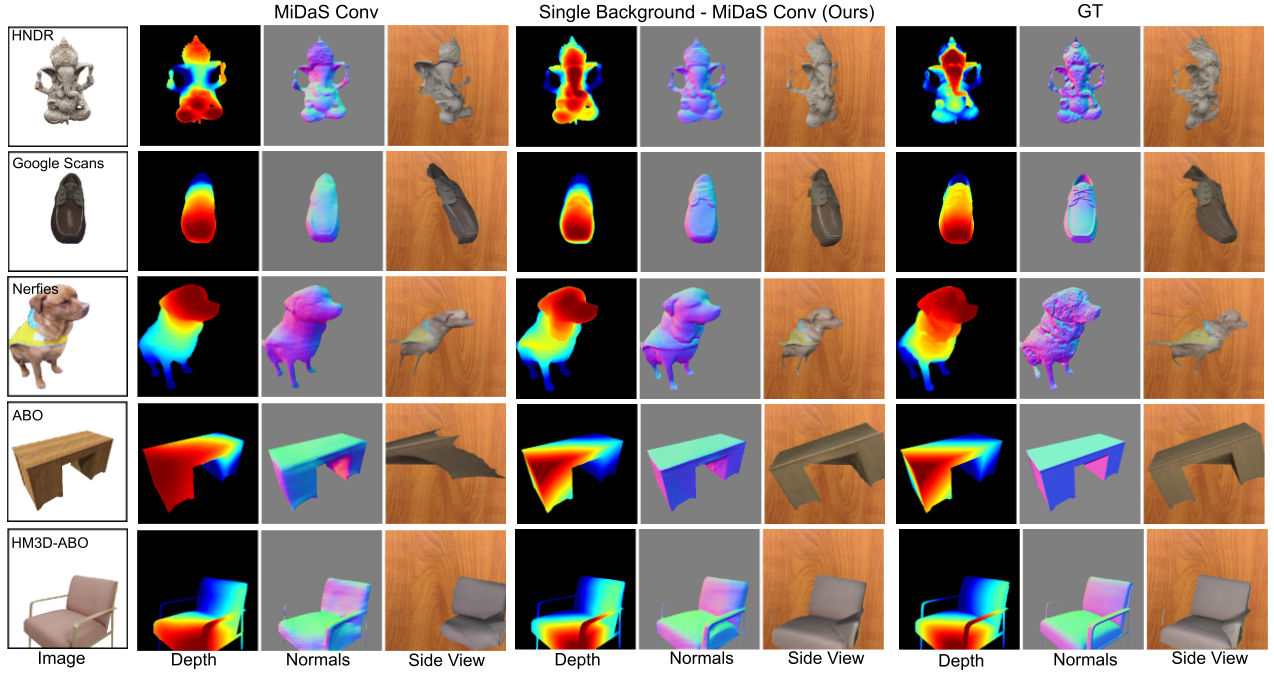}
\end{center}
\vspace{-0.20in}
\caption{\textbf{Qualitative results for MiDaS Convolutional} Depth, normals, and mesh (viewed from the side) for samples of each of the testing datasets, for MiDaS Convolutional before and after adaptation with a single background. }
\label{fig:results_midas_conv_single_background}
\vspace{-0.1in}
\end{figure*}

\begin{figure*}[h]
\begin{center}
\includegraphics[width=1\linewidth]{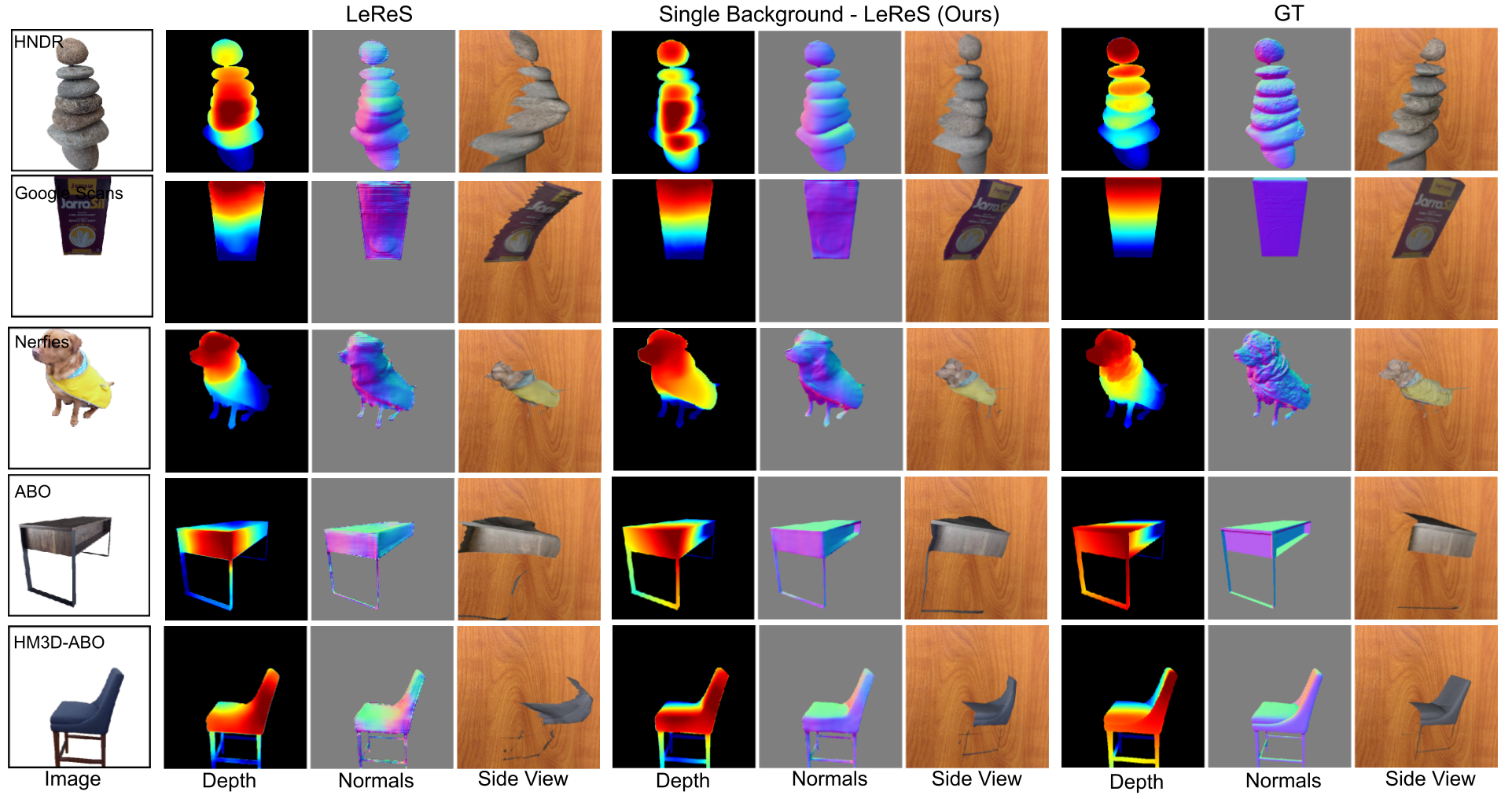}
\end{center}
\vspace{-0.20in}
\caption{\textbf{Qualitative results for LeReS} Depth, normals, and mesh (viewed from the side) for samples of each of the testing datasets, for LeReS before and after adaptation with a single background. }
\label{fig:results_leres_single_background}
\vspace{-0.1in}
\end{figure*}

\begin{figure*}[h]
\begin{center}
\includegraphics[width=1\linewidth]{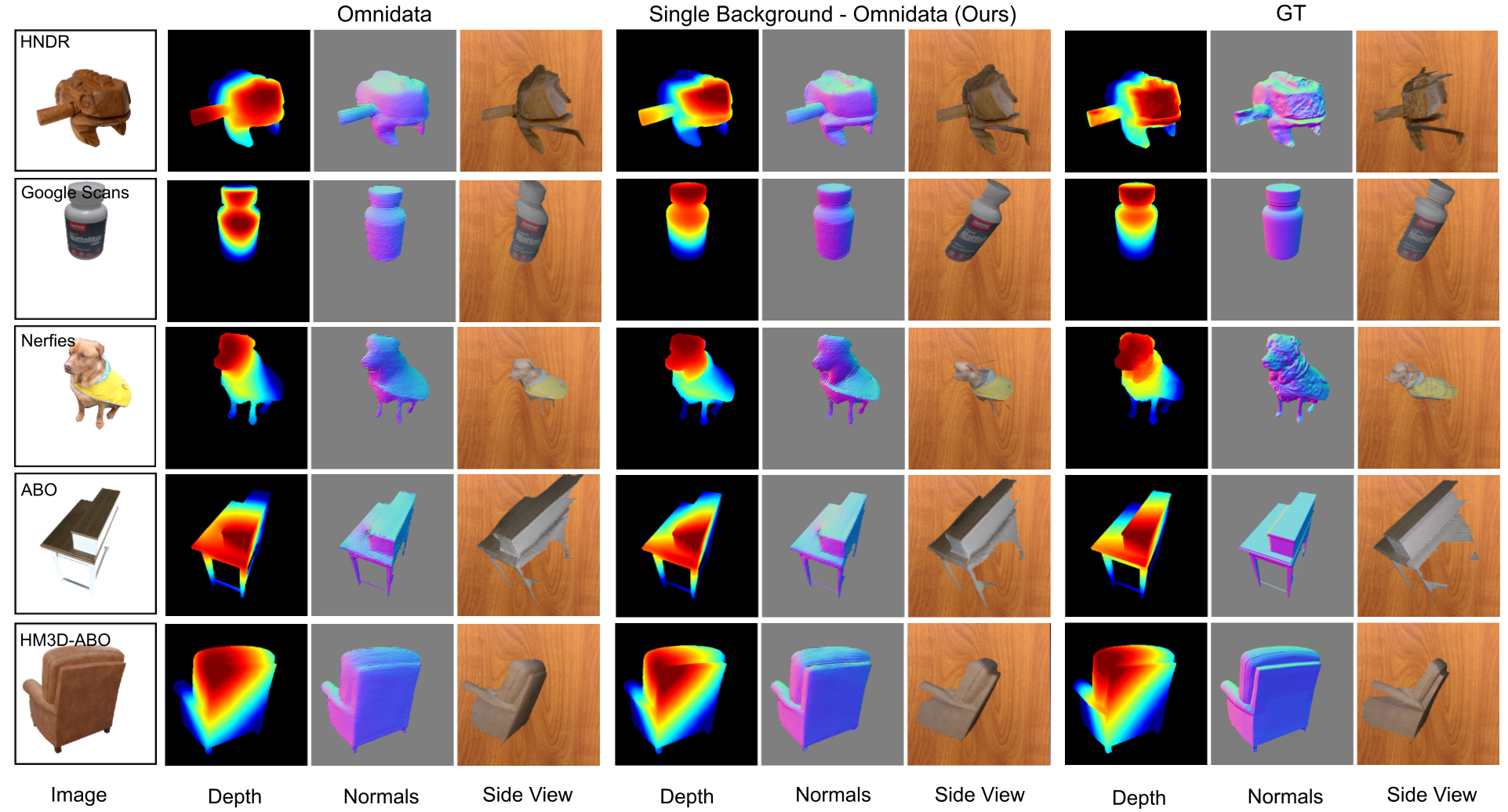}
\end{center}
\vspace{-0.20in}
\caption{\textbf{Qualitative results for Omnidata} Depth, normals, and mesh (viewed from the side) for samples of each of the testing datasets, for Omnidata before and after adaptation with a single background. }
\label{fig:results_omnidata_single_background}
\vspace{-0.1in}
\end{figure*}

\begin{figure*}[h]
\begin{center}
\includegraphics[width=1\linewidth]{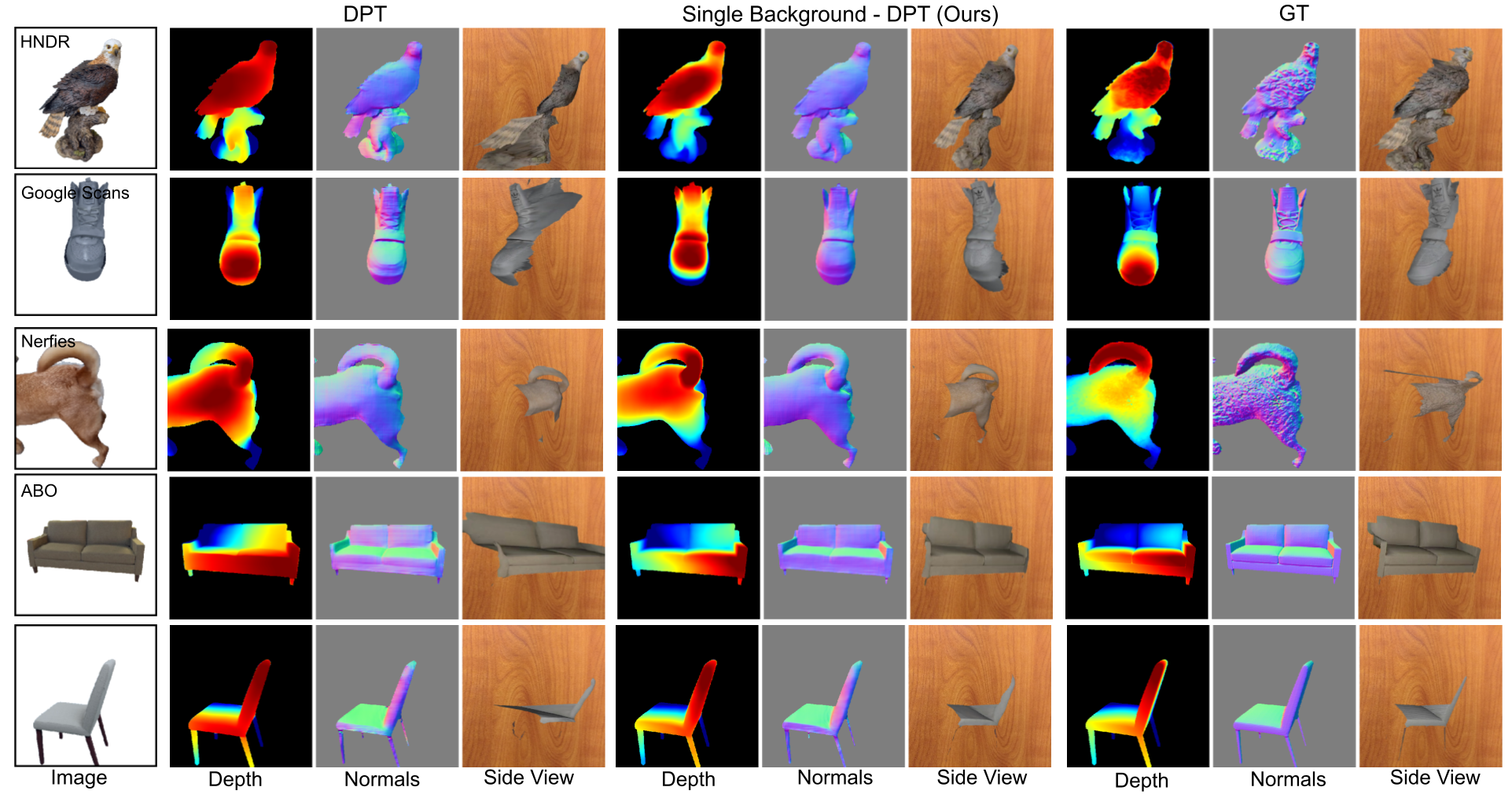}
\end{center}
\vspace{-0.20in}
\caption{\textbf{Qualitative results for DPT} Depth, normals, and mesh (viewed from the side) for samples of each of the testing datasets, for DPT before and after adaptation with a single background. }
\label{fig:results_dpt_single_background}
\vspace{-0.1in}
\end{figure*}

\begin{figure*}[h]
\begin{center}
\includegraphics[width=1\linewidth]{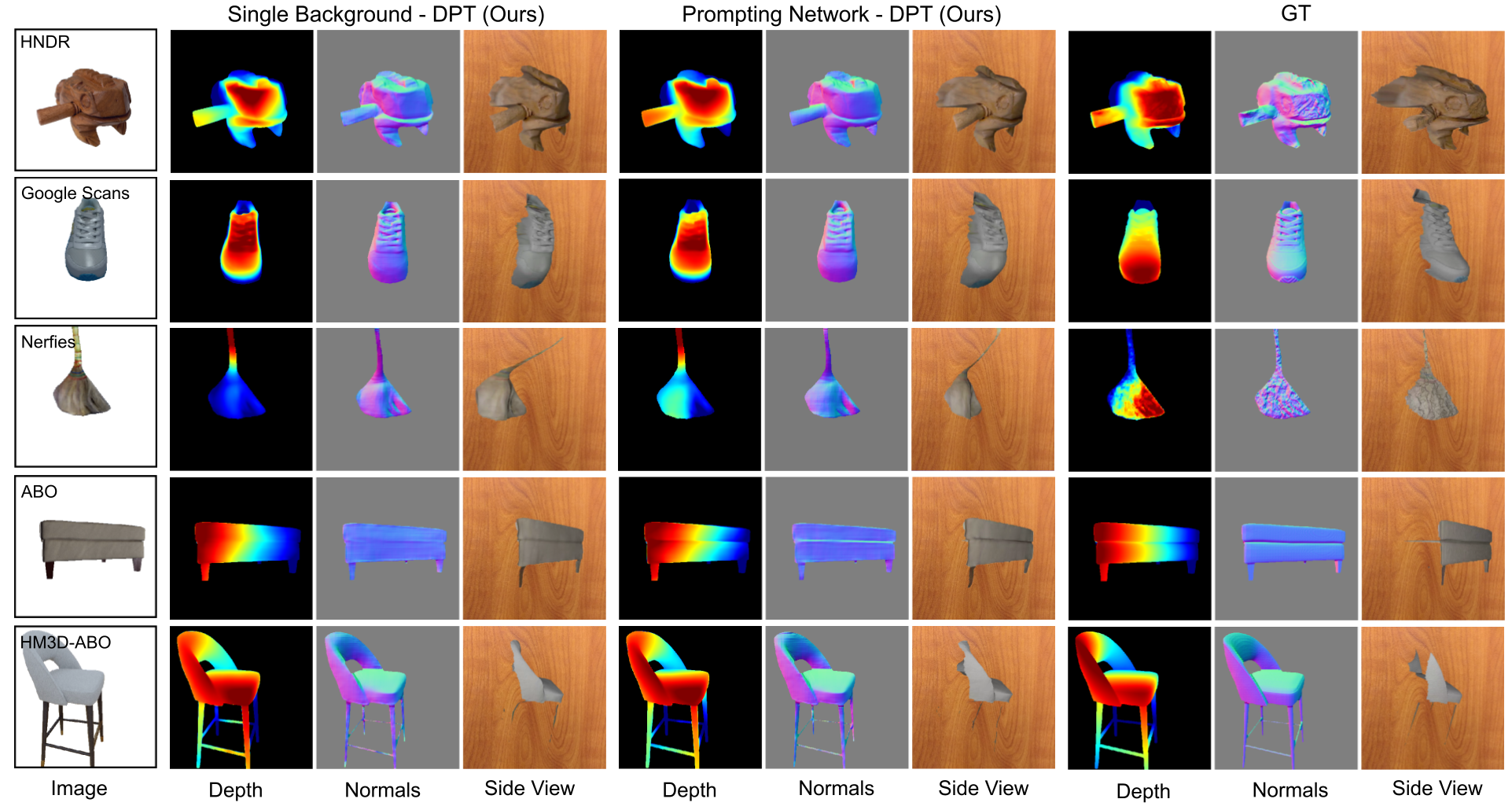}
\end{center}
\vspace{-0.20in}
\caption{\textbf{Single Background vs Prompting Net} Depth, normals, and mesh (viewed from the side) for samples of each of the testing datasets, for DPT with a Single Background and the Prompting Network strategy. }
\label{fig:single_background_vs_prompting_network}
\vspace{-0.1in}
\end{figure*}

In Figures~\ref{fig:results_midas_conv_single_background}-\ref{fig:results_dpt_single_background}, we show qualitative results for our Single Background method against the original baselines. In Figure~\ref{fig:single_background_vs_prompting_network} we show selected results comparing the single background strategy against the Prompting network strategy, to highlight the cases where the Prompting network outperforms the single background.

